# Kids Computer Interaction with a Focus on Story Telling and Educational Robotics

Cristina Gena
Computer Science Department, University of Turin
corso Svizzera 185, 10149, Turin
cristina.gena@unito.it

**Abstract**

In the invited talk at KDAH 2018 workshop (CIKM 2018 conference) I have presented two projects carried out in our HCI lab and target to child: the first one is related to the evaluation in the wild of of a storytelling app for kids based on the swipe story approach, the second one is related to Wolly, an educational robot co-designed with children.

## 1 Introduction

Child-computer interaction concerns the study of the design, evaluation, and implementation of interactive computer systems for children, and the wider impact of technology on children and society. According to Hourcade [4], this definition, paraphrasing the Association for Computing Machinerys (ACM) definition of human-computer interaction, lists design, evaluation and implementation in an order in which they normally do not occur. This is intentional, as most human child-computer interaction research is about design, followed by evaluation, followed by implementation.

In the invited talk I have presented two projects carried out in our HCI lab, which will be described in the following.

## 2 Story Telling for Kids

In this section I report the preliminary results of an evaluation in the wild of the storytelling app (for more details see [3]), the Lions of Time[1], presenting an archeological story, based on the swipe story approach[2]. A swipe story is a digital story that, with a simple and immediate gesture (namely the swipe) and a language based on drawings, images, words, games, sounds, movies and emotions, is able to start the young user to a new path of knowledge. The swipe story was launched at the 2017 Turin International Book Fair[3]. The swipe story is realized throughout a tablet and smartphone app, freely available in Google Play and App store markets[4]. This swipe story supports an illustrated novel [1], and digital storytelling has been applied with the aim to bring the young audiences closer to the knowledge of the cultural heritage of Friuli Venezia Giulia, the Italian region situated further to the North East.

The story takes the user to discover the archaeological sites of Friuli Venezia Giulia, where the present is intertwined with the past, reality with imagination. Swipe after swipe, the user is walking in space-time in the company of the three characters (Eleonora, Leonardo, and Ruggero) between reconstructions of places, cities (Pradis and Aquileia) and ancient monuments, on the light wings of fantasy.

During the 2017 Turin International Book Fair, Le Muse Archaeological Association[5] organized a public event at the Fruili Venezia Giulia stand, for presenting to the public the swipe story The Lions of Time. For collecting an initial and spontaneous feedback from real users, we invited to the event a group of families having children in the 7-9 age range. The children who participated at the event were 12, 5 females and 7 males. 9 of them have extensively played the swipe story, and we have based our main observations on their interactions. For the trial, we had 7 Android-based tablets and 2 Android-based smartphones. We left the children free to interact with the swipe-story, giving as little instructions as possible, telling them only that they were there for the launch of a new app for children, and they would be the first to try it. In fact we were there in 4 observing them, and one of us video-recorded the children and their interactions for further post hoc analysis.

---

[1] *I Leoni del Tempo - Archeostorie del Friuli Venezia Giulia - The Lions of Time - Archeo-stories of Friuli Venezia Giulia* is a trans-media editorial project promoted by ERPAC (Regional Institution for the Cultural Heritage of the Autonomous Region of Friuli Venezia Giulia, Cataloging, Training and Research Service)

[2] http://www.swipe-story.com/
[3] https://www.salonelibro.it/
[4] http://www.swipe-story.com/app/ileonideltempo
[5] http://www.lemusestudio.it/home.html

This limited and initial evaluation in the wild (inspired by the methodology described in [5]) showed how the mobile app has been specially developed in order to win the attention of the little ones. The children, despite being very young, have not had any problem in the immediate use of the story, a sign that the digital natives are used to using new technologies not only to play, but also to communicate and above all to learn. In the short period of use they managed to quickly learn what the app showed, even if with important and culturally advanced contents, without getting bored or learning difficulties. As future improvement, we should think of a series of images with comics and/or narrating voice also for the in-depth analysis provided when clicking on question marks, in order to improve the level of the engagement proposed by the app. Another improvement could be that of expanding the quantity of games proposed at the end, also linking them to moments of didactical learning and deepening, and introducing gamification mechanisms to increase the children involvement.

## 3 Educational Robotics

In our HCI lab, we carried out a co-design activity with children aimed at devising an educational robot called Wolly (for more details see [2]). The main goal of the robot is acting as an affective peer for children: hence, it has to be able to execute a standard set of commands, compatible with those used in coding, but also to interact both verbally and affectively with students. We are now working on controlling Wolly by means of a standard visual block environment, Blockly[6], which is well know to many children with some experience in coding. However, we would also like to have a simpler set of instructions, specifically designed for Wolly, so that children can use basic commands to control its behavior.

As far as the co-design is concerned, in November 2017, we carried out a co-design session with 25 children, described in detail in [2]. All children were in the third grade, 8 to 9 years old, and with no experience in educational robotics. Following a co-design methodology, they were asked to provide suggestions for some features of the robot: its name, physical appearance, facial expressions, personality and character. Based on the insights drawn from the co-design process, we designed the robot appearance and structure. In particular, the robot - built using a common hobby robotic kit - is able to move through its four independent motorized wheels and can be controlled through either a web application or a set of Android apps that contact its REST APIs. Its body has been almost completely 3D printed, while its head consists in an Android-based smartphone able to show and perceive emotions, to produce verbal expressions and to understand voice commands.

As far as interactive features are concerned, Wolly plays the role of an educational robot that helps kids in coding exercises, giving them suggestions on how to reach their goals and write their code, at the same time being able to execute instructions such as moving on a chessboard, as other educational robots can do. However, since Wolly is also able to interact with kids in a verbal and affective way, we would like to enable children to program its basic behaviors and social interactions, in order to teach them the basis of social robot programming.

---

[6]https://developers.google.com/blockly/